\documentclass[conference]{IEEEtran}
\IEEEoverridecommandlockouts
\usepackage{cite}
\usepackage{amsmath,amssymb,amsfonts}
\usepackage{algorithmic}
\usepackage{graphicx}
\usepackage{textcomp}
\usepackage{xcolor}
\usepackage{multirow}
\usepackage{booktabs}
\usepackage{hyperref}
\usepackage{authblk}
\usepackage{tikz}
\usepackage{subcaption}
\usepackage[font=small,labelfont=bf]{caption}
\newcommand*\circled[1]{\tikz[baseline=(char.base)]{
            \node[shape=circle,draw,inner sep=1pt] (char) {#1};}}
\hypersetup{
    colorlinks=true,
    linkcolor=blue,
    filecolor=magenta,      
    urlcolor=cyan,
    citecolor = magenta,
    pdfpagemode=FullScreen,
}

\def\BibTeX{{\rm B\kern-.05em{\sc i\kern-.025em b}\kern-.08em
    T\kern-.1667em\lower.7ex\hbox{E}\kern-.125emX}}
\begin{document}

\title{
Robotic Computing on FPGAs: Current Progress, Research Challenges, and Opportunities
\vspace{-5pt}

}
\author[$1$]{Zishen~Wan}
\author[$1$]{Ashwin~Lele}
\author[$2$]{Bo~Yu}
\author[$2$]{Shaoshan~Liu}
\author[$3$]{Yu~Wang}
\author[$4$]{\\Vijay~Janapa~Reddi} 
\author[$1$]{Cong~Hao} 
\author[$1$]{Arijit~Raychowdhury\vspace{-5pt}} 

\affil[$1$]{\normalsize{ School of Electrical and Computer Engineering, Georgia Institute of Technology, Atlanta, GA, USA}}
\affil[$2$]{ PerceptIn, Fremont, CA, USA}
\affil[$3$]{ Department of Electronic Engineering, Tsinghua University, Beijing, China}
\affil[$4$]{ School of Engineering and Applied Sciences, Harvard University, Cambridge, MA, USA}
\affil[$ $]{\textit{\{zishenwan, alele9, callie.hao\}@gatech.edu, 
arijit.raychowdhury@ece.gatech.edu}} 
\affil[$ $]{\textit{\{bo.yu, shaoshan.liu\}@perceptin.io, yu-wang@tsinghua.edu.cn, vj@eecs.harvard.edu \vspace{-10pt}
}}

\maketitle

\begin{abstract}
Robotic computing has reached a tipping point, with a myriad of robots (e.g., drones, self-driving cars, logistic robots) being widely applied in diverse scenarios. The continuous proliferation of robotics, however, critically depends on efficient computing substrates, driven by real-time requirements, robotic size-weight-and-power constraints, cybersecurity considerations, and dynamically changing scenarios. Within all platforms, FPGA is able to deliver both software and hardware solutions with low power, high performance, reconfigurability, reliability, and adaptivity characteristics, serving as the promising computing substrate for robotic applications. This paper highlights the current progress, design techniques, challenges, and open research challenges in the domain of robotic computing on FPGAs.

\end{abstract}

\section{Introduction}
\label{sec:intro}
Robotic computing is on the rise. A myriad of robots such as drones, legged robots, and self-driving cars are on the verge of becoming an integral part of our life~\cite{wan2021survey,liu2021robotic}. Robotics is typically an art of system integration both in software and hardware (Fig.~\ref{fig:overview}). The continuous proliferation of robots, however, face computing challenges, raised from the higher performance requirements, resource constraints, miniaturization of machine form factors, dynamic operating scenarios, and cybersecurity considerations. Therefore, it is essential to choose a proper computing substrate for robotic system that can meet real-time and power requirements and adapt to changing workloads.

CPUs and GPUs are two widely-used computing platforms, however, their performance and efficiency are still incompetent in real-time computation for complex robots. 
Take the motion planning task as an example, CPU typically takes a few seconds to find the collision-free trajectory~\cite{hauser2015lazy}, making it too slow for complex navigation tasks. GPUs can finish planning tasks in hundreds of milliseconds, still insufficient for many scenarios while at hundreds of watts cost~\cite{pan2012gpu}. ASICs are recently developed for specific robotic workloads with low power and high performance~\cite{suleiman2019navion,yoon2020neuroslam,wan2022circuit}, but their fixed architecture has difficulty in adapting to rapid-evolving robotic algorithms and dynamic scenarios, and is vulnerable to cybersecurity threats.

As an alternative, we believe FPGA is the promising compute substrate for robotic applications. First, FPGA increases the performance with massive parallelism and deeply pipelined datapath, making it capable of meeting real-time requirements with high energy efficiency compared to CPUs and GPUs. Second, FPGA can adaptively generate custom architectures and update with the fast-evolving of robotic algorithms without going through re-fabrication as ASIC~\cite{mayoral2021adaptive}. Third, FPGA is flexible in dealing with highly diverse robotic workloads, especially with partial reconfiguration allowing modification part of the operating board. Fourth, FPGA provides reliable design by leveraging reconfiguration to patch flows, compared to potential vulnerabilities detected in fixed architectures~\cite{kocher2019spectre}, which is especially essential in safety-critical scenarios~\cite{wan2021analyzing}. Overall, FPGA has the potential to deliver high-performance, low-power, reconfigurable, adaptive, and secure features in robotic computing, and is booming in autonomous applications. However, several challenges, such as tedious development procedures, inefficient system support, and huge design space, remain in the FPGA-based robotic computing and impede the way ahead.

In this paper, we will discuss the current progress, challenges, and opportunities for FPGA-based robotic computing. Section~\ref{sec:algo} introduces the cross-layer stack of robotic system. Section~\ref{sec:FPGA} presents current FPGA accelerators and systems for robotic computing, with an emphasis on design techniques. Section~\ref{sec:challenges} discusses challenges and opportunities for FPGA-based robotic computing, and our view of the road ahead.

\section{Cross-Layer Robotic Computing Systems}
\label{sec:algo}
\begin{figure}[t!]
        \centering\includegraphics[width=\columnwidth]{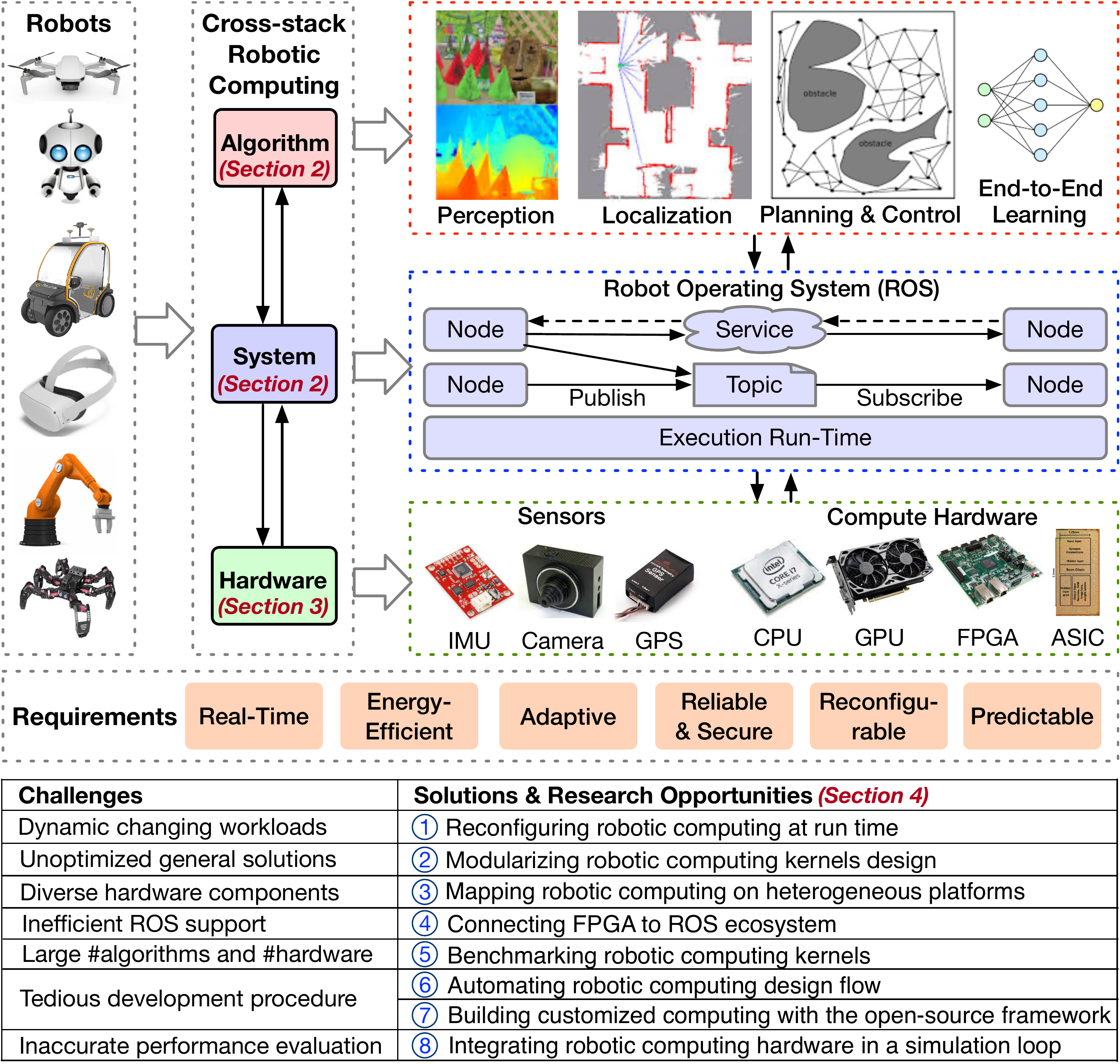}
        \caption{Cross-layer stack of the robotic computing system, requirements, research challenges, and open opportunities.}
        \label{fig:overview}
        \vspace{-10pt}
\end{figure}

This section introduces the abstraction layers of the robotic computing stack. We traverse down Fig.~\ref{fig:overview} to explain robotic-specific algorithms and systems building blocks.

\subsection{Robotic-Computing Algorithm Layer}
\begin{figure}[t!]
        \centering\includegraphics[width=\columnwidth]{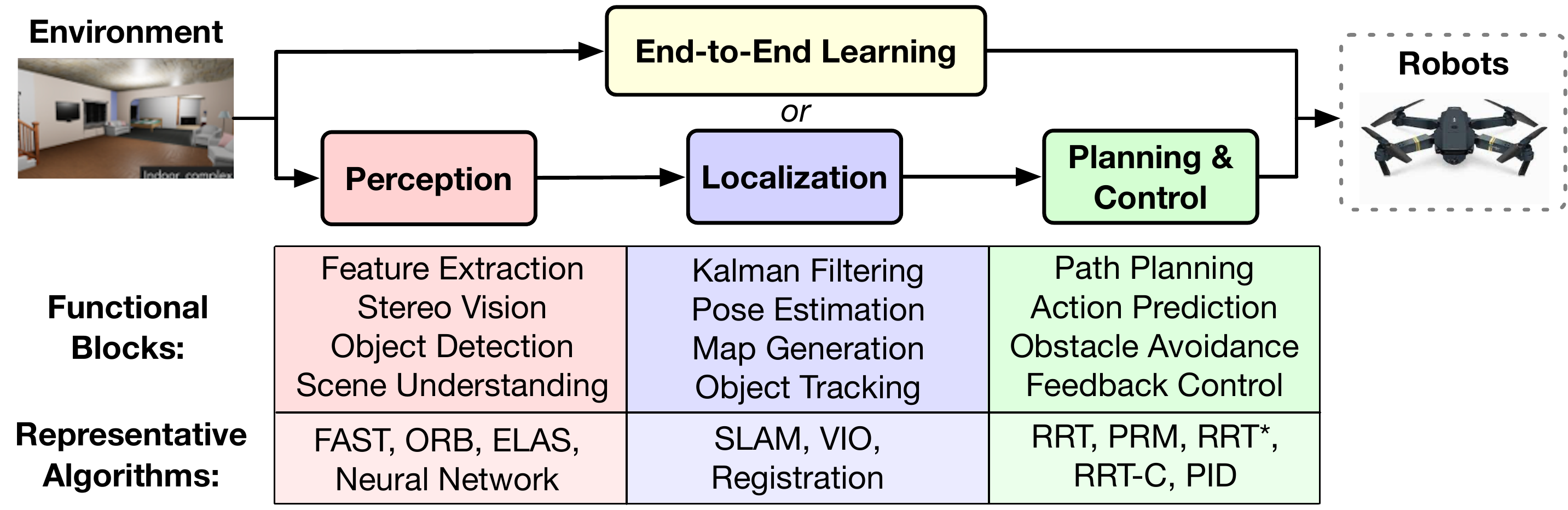}
        \caption{Applications and algorithm building blocks in robotic systems.}
        \label{fig:algorithm}
        \vspace{-10pt}
\end{figure}

Fig.~\ref{fig:algorithm} illustrates the representative algorithm building blocks in robotic computing, including sense-plan-act (perception, localization, planning, control) and end-to-end learning.

\textbf{Perception.}
The goal of perception is to sense the dynamic surroundings and build a reliable and detailed representation based on sensory data (e.g., camera, IMU, GPS, LiDAR). Perception usually includes feature extraction, stereo vision, object detection, scene understanding, etc. In feature extraction, key points are usually detected using FAST feature and ORB descriptor~\cite{rublee2011orb}. Compared with all image pixels, operating on feature points can improve the robustness and compute efficiency. Stereo vision is to obtain 3D structure information of the scene through disparity calculation. Local, semi-global, and global stereo matching algorithms are proposed based on operational scenarios~\cite{lu2021resource}. Recently, advances in deep learning have exposed robotic perception systems to more tasks.

\textbf{Localization.}
The goal of localization is to calculate the position and orientation of a robot itself in a given frame of reference. Knowing the position fundamentally enables robots to plan the trajectory and navigate, and knowing the orientation further helps robots stabilize. Simultaneous localization and mapping (SLAM) is a commonly-used algorithm where the robot simultaneously constructs a map of the environment while localizing itself~\cite{mur2017orb}, and one principled mathematical approach to solving SLAM is maximum a posteriori estimation. The filtering-based approach has recently been developed with Multi-State Kalman Filter-based algorithms such as MSCKF VIO~\cite{sun2018robust} and OpenVINS~\cite{geneva2020openvins}.

\textbf{Motion planning and control.}
The goal of motion planning is to find the optimal collision-free trajectory from the start position to the goal position, which is invoked during a robot movement to adapt to environmental changes. Motion planning is usually followed by a control module continuously tracking the differences between actual poses and poses on the pre-defined trajectory. Sampling-based solutions are widely used for motion planning, such as Probabilistic Roadmap (PRM)~\cite{ichter2020learned}, Rapidly-exploring Random Tree (RRT)~\cite{lavalle2001rapidly} and their variants, which generally contain three steps: roadmap construction, collision detection, and graph search.

\textbf{End-to-End learning system.} 
End-to-end algorithms enable skill learning directly from sensor input and perform all the following cognitive robotic tasks using a single neural network model. Maps or separate planning stages are not required in end-to-end learning. The neural network model can be trained using reinforcement learning~\cite{anwar2020autonomous} or supervised learning~\cite{loquercio2018dronet}. The challenges of end-to-end learning include alleviating the model simulation-to-reality performance gap, designing optimal reward functions, and improving model explainability and robustness, which are actively explored.

\subsection{Robotic-Computing System Layer}
\textbf{Robot Operating System (ROS).}
ROS is a commonly used operating system to provide tools, libraries, and package management for robotics development.
It is a distributed framework of processes that enables executables to be individually designed and loosely coupled at runtime. 
Conceptually, the peer-to-peer network of ROS processes is called computation graph. The basic ROS computation graph includes nodes, topics, services, and masters, all of which provide data to the graph in different ways (Fig.~\ref{fig:overview}). Each ROS node is a process used to perform a task. ROS nodes communicate with each other via topics or services. Topics allow one node to publish messages that multiple other nodes can subscribe. Services allow for creating a one-to-one communication between a service node and a client node.
The ROS master is responsible for storing operating parameters and managing other nodes.

\section{Current Progress and Design Techniques}
\label{sec:FPGA}
This section presents our designs and current progress for FPGA-based robotic computing, with an emphasis on the design traits and techniques.

\textbf{Perception on FPGAs.}
The perception typically contributes significantly to the end-to-end latency of robotic applications. Take the ORB perception module as an example, it usually accounts for 50\%-80\% compute latency of the whole localization scenario. To alleviate that, \cite{wan2021energy,gan2021eudoxus} accelerate ORB-based perception on FPGA for both aerial and ground robots. The key design principles are to exploit task-level parallelisms by frame-multiplexing feature extraction, and customize on-chip memories to suit different types of data reuse. Another series of illustrative works is to accelerate stereo matching, which is the bottleneck of the stereo vision system. \cite{perri2020stereo} implements local stereo matching algorithms on FPGA with characterized hardware-software partitions. \cite{kamasaka2018fpga} proposes a parallel 3D graph cut algorithm for accelerating global stereo matching, achieving 166$\times$ speedup to CPU. FP-Stereo~\cite{zhao2020fp} present streaming architecture and sampling-insensitive disparity algorithm on FPGA to accelerate semi-global stereo matching. Recently, the Bayesian approach with generative probabilistic models facilitates efficient dense matching. An appealing example is iELAS~\cite{gao2021ielas}, a hardware-friendly large-scale stereo algorithm implemented on FPGA. iELAS reforms the computational-intensive and irregular triangulation modules in a regular manner with intelligent points interpolation. Additionally, FPGA has been widely used in accelerating neural networks for robotic perception and end-to-end learning. Several techniques, including quantization, loop optimization, array partitioning, data reuse, and memory optimization, are proposed. Interested readers are pointed to~\cite{abdelouahab2018accelerating} for details.

\textbf{Localization on FPGAs.}
The backbone of SLAM is a complex non-linear optimization problem, bundle adjustment (BA), which consumes a significant amount of time and power. $\pi$-BA~\cite{liu2020pi} designs a co-observation optimization technique to accelerate BA based on the key inspection that not all 3D points appear on all images in a BA problem. Pisces~\cite{asgari2020pisces} co-optimizes SLAM power consumption and latency by exploiting inherent SLAM sparsity. By orchestrating sparse data, Pisces aligns correlated data and enables direct, parallel, and deterministic memory access. Going beyond point solutions, Archytas~\cite{liu2021archytas} presents a template hardware synthesis solution that automatically generates a SLAM accelerator given the hardware template and algorithm data-flow graph. To make the design adaptable to various environments, \cite{liu2022energy} dynamically optimizes the SLAM accelerator with an offline constructed lookup table and clock gating without sending new bitstreams to FPGA at run time. Typically, SLAM is suitable for unknown indoor environments, while Registration is used for known indoor environments and visual-inertial-odometry (VIO) functions well for outdoor environments. No single localization algorithm fits all scenarios. Interestingly, these algorithms share fundamental computation kernels amenable to matrix blocking. Eudoxus~\cite{gan2021eudoxus} implements SLAM, Registration, and VIO on FPGA by accelerating common matrix operations, with a lightweight runtime scheduler reducing variation.

\textbf{Motion planning and control on FPGAs.}
Among the motion planning pipeline, the computation of collision detection is usually the bottleneck. Take RRT as an example, when it runs on CPU, 99\% of the instructions are executed for collision detection, taking up 90\% of total computation time. Recent efforts have proposed to accelerate motion planning kernels through algorithm-hardware co-design on FPGAs. \cite{murray2016microarchitecture} constructs robot-specific circuitry and architecture with roadmap pre-computation and massive path search parallelism, which is able to solve a motion planning query in 16~$\mu$s. \cite{murray2019programmable} further presents a programmable dataflow architecture with a low-cost interconnection network, reducing the latency to 2.3~$\mu$s. 

Several key design and optimization techniques are leveraged in FPGA-accelerated perception, localization, planning, and control. From the software aspect, hardware-friendly algorithms and data structures are proposed to promote parallelism with reduced intrinsic recursions and algorithm complexity. From the hardware aspect, robot-specific architecture, data sparsity, locality, parallelism, optimized interconnection networks, and reduced data movement contribute to high-performant and flexible motion planning design. These techniques can be generalized to other implementations, serving as a guide for future works.

\textbf{Multi-robot collaboration on FPGAs.}
Going beyond single-robot applications, swarm robotics has been increasingly deployed in real-life scenarios where a team of robots collaboratively finish a task. Multi-robot workload typically demonstrates unique compute challenges. Several algorithm kernels may need to process the data at the same time, leading to hardware resources conflicts. Therefore, the FPGA accelerator should support multi-thread and dynamic scheduling. An intriguing example is INCAME~\cite{yu2021incame}, a single-core multi-robot exploration framework that supports dynamic multi-task scheduling with a virtual-instruction-based interrupt method. The perception and control tasks are assigned high priorities, while long-term decisions and optimization have low priorities. We envision that the multi-core multi-tasking FPGA accelerator will further improve the performance of the multi-robot system.

\textbf{ROS on FPGAs.}
ROS-compliant FPGAs have recently been developed with ROS becoming increasingly common in robotics. ROS-compliant FPGAs must consider four functions: encapsulation of FPGA circuits, interface between ROS software and FPGA hardware, subscribe interface, and publish interface to ROS topic. Typically, large communication latency between ROS components is the bottleneck of offloading computing to FPGAs. \cite{sugata2017acceleration} reduces the latency by implementing publish and subscribe messaging of ROS as hardware circuits, making the direct ROS-FPGA communication possible and efficient. Recently, \cite{leal2020automated,lienen2020reconros} propose tools and frameworks to offload and accelerate ROS computational graph on FPGA. However, the ecosystem of ROS on FPGAs is still in its infancy, better interface, automated tools, and whole ROS acceleration are to be developed.

\section{Research Challenges and Future Directions}
\label{sec:challenges}
This section discusses the research opportunities for FPGA-based robotic computing, and our view for road ahead (Fig.~\ref{fig:overview}).

\circled{1} \textbf{Reconfiguring robotic computing at run time.} 
Robots usually operate in highly dynamic environments, thus designing runtime-reconfigurable compute platforms is critical and can enable robots to be adaptive in various scenarios.
Partial reconfiguration (PR) is a key feature of FPGA. Using PR, part of FPGA can be reconfigured at runtime without compromising the integrity of the applications running on those parts of the device that are not being reconfigured. 
Therefore, PR can allow various robotic computing kernels to time-share part of an FPGA, leading to high performance and energy efficiency, and making FPGA a more suitable computing platform for dynamic and complex robotic workloads. 

\circled{2} \textbf{Modularizing robotic computing kernels design.} 
The number of robotic algorithms is booming, but many algorithm variants share similar key computation blocks. It is thereby imperative to modularize the robotic computing kernel design.
We can build optimized hardware acceleration blocks for these kernels as libraries or packages, while exploring their inherent task-specific features such as sparsity, data flow, and memory access patterns. During the design phase, robotics practitioners can directly import these robotics-specific libraries and building blocks to build their FPGA design without delving into hardware engineering, which will greatly ease the design process.
Modularizing the robotic algorithm design can help roboticists create custom accelerators for a kernel without hardware expertise.

\circled{3} \textbf{Mapping robotic computing on heterogeneous platforms.} 
One of the key technical challenges of designing robotic compute systems is to develop a suitable computer architecture, along with a software stack that allows computational flexibility.
To improve the overall performance, FPGA-based System-on-Chip (SoC) solutions for robotic computing would be of the essence~\cite{mayoral2021adaptive}, which holistically integrates various computing technologies, including CPU, GPU, FPGA, and accelerators.
The OpenCL framework can be used for programming and executing programs across heterogeneous platforms, and accelerator-level parallelism is expected to be explored~\cite{hill2021accelerator}.
By doing so, the SoCs are equipped with both software and hardware programmability, having the capability to deliver high performance, low power, adaptive, and reliable robotic computing.

\circled{4} \textbf{Connecting FPGA to ROS ecosystem.} 
With ROS increasingly utilized in robotics applications of all scales, robotic FPGA platforms need to be able to efficiently map ROS computational graphs on silicon.
Going beyond the current work on accelerating specific ROS libraries, the inter-process and intra-process between ROS nodes also need to be accelerated~\cite{lienen2022reconros}. It is worth noting that the hardware acceleration must be directly integrated into the ROS ecosystem to provide a seamless user experience for roboticists. A better interface between ROS and FPGA is expected to be delivered.
Furthermore, through dynamically and efficiently mapping ROS to heterogeneous compute platforms, holistic hardware acceleration for robotic computing on ROS applications is expected to be achieved.

\circled{5} \textbf{Benchmarking robotic computing kernels.} 
Given the proliferation of robotic kernels and the rapid advances of hardware platforms, benchmarking these robotic algorithms and systems in a comparable, quantitative, and validatable manner is imperative.
Such benchmarking comes into two folds, benchmarking a robotic algorithm across various hardware platforms, and benchmarking various robotic algorithms within the same hardware~\cite{neuman2019benchmarking}. Particularly, benchmarks should consider the interactions of ROS and its computational graph. 
Benchmarking robotic computing will guide the robotics and hardware researchers to investigate the trade-offs in accuracy, performance, and energy efficiency of various robotic algorithms, and implement (or select) algorithms on FPGAs and other platforms in a performance-portable way.

\circled{6} \textbf{Automating robotic computing design flow.} 
Given the increasing complexity of robotic algorithms and the cross-stack nature, the development of robotic computing systems is becoming slow and tedious.
Thus, building a push-button flow with robotic task requirements as input to automatically generate robotic accelerator design is critical~\cite{krishnan2021autopilot,neuman2021robomorphic,krishnan2021autosoc}.
We envision the agile framework will intelligently search the huge design space and automatically choose the optimal algorithm-hardware parameters with the help of modular kernels, benchmarking, and machine learning-assist methods. New robotic-centric electronic design automation (EDA) tool needs to be developed to convert the design to FPGA implementation.
Automating the design follow will greatly facilitate the FPGA-based robotic computing development, and make FPGAs an ideal platform for fast prototyping and commercialization.

\circled{7} \textbf{Building customized robotic computing with the open-source framework.} 
The field of robotic computing is still in its infancy and fast-changing, and numerous opportunities still exist in task-specific acceleration. 
The open-source design framework with iteratively deployment, profiling, and optimization has recently been developed for machine learning applications~\cite{prakash2022cfu}, but it is still to be explored for robotic computing applications. Designers can build their custom specialized and optimized processors based on the RISC-V instruction set architecture (ISA).
Defining and building an open-source FPGA-based RISC-V robotics-on-chip processor with open-source frameworks would considerably facilitate the design process and allow us to adapt to the rapidly changing landscape of robotic computing algorithms and accelerators.

\circled{8} \textbf{Integrating robotic computing hardware in a simulation loop.}
The FPGA-accelerated kernels are usually part of the whole autonomy computing pipeline. The correlation among compute stages and other robotic cyber-physical components will impact the final robotic system performance and lead to inaccurate hardware evaluation~\cite{krishnan2020sky,krishnan2022roofline}. Thus, instead of isolated hardware development, adopting the hardware-in-the-loop (HIL) method is critical~\cite{boroujerdian2018mavbench}. HIL requires plugging the hardware platforms into the simulation to understand how robots respond to stimuli on FPGA or other compute substrates. HIL can help designers quantify the FPGA real-time performance within the whole system and enable robust evaluation without risking real robots. Particularly, HIL can alleviate the FPGA hardware-induced gaps between training and deployment in learning-based systems. To perform faster performance evaluation at an earlier design stage, a closed-loop co-simulation framework of both FPGA architectural behavior (e.g., FireSim~\cite{karandikar2018firesim}, SystemModeler~\cite{acevedo2016fpga}) and robotic environment simulator (e.g., AirSim~\cite{shah2018airsim}) is necessary.

The abundance of challenges raised above provides plentiful opportunities for research development at all levels. Endeavoring to solve these problems requires interdisciplinary approaches across all layers of computing stack, from algorithm and system to architecture, micro-architecture, and circuits.

\section{Conclusion}
\label{sec:conclusion}
Robotic computing is a rising area and critically depends on efficient, adaptive, and reliable compute substrates. This paper presents the cross-layer robotic computing stack and illustrates the current progress, along with FPGA design techniques.
We conclude the paper by discussing the challenges, research opportunities, and roadmap for the next-generation FPGA-based robotic computing systems.

\bibliographystyle{ieeetr}
\bibliography{refs}

\end{document}